\documentclass[conference]{IEEEtran}
\IEEEoverridecommandlockouts
\usepackage{times}
\usepackage{epsfig}
\usepackage{graphicx}
\usepackage{subfigure}
\usepackage{amsmath}
\usepackage{amssymb}
\usepackage{cite}
\usepackage{amsfonts}
\usepackage{algorithmic}
\usepackage{textcomp}
\usepackage{xcolor}
\usepackage{array}
\usepackage{multirow}
\usepackage{bbm}
\usepackage[pagebackref=true,breaklinks=true,colorlinks,bookmarks=false]{hyperref}

\def\BibTeX{{\rm B\kern-.05em{\sc i\kern-.025em b}\kern-.08em
    T\kern-.1667em\lower.7ex\hbox{E}\kern-.125emX}}
\begin{document}

\title{Improving Medical Image Classification with Label Noise Using Dual-uncertainty Estimation}

\author{Lie Ju \\
\IEEEauthorblockA{Airdoc \\
Monash University\\
julie334600@gmail.com}
\and
\IEEEauthorblockN{Xin Wang}
\IEEEauthorblockA{Airdoc \\
wangxin@airdoc.com}
\and
\IEEEauthorblockN{Lin Wang}
\IEEEauthorblockA{Airdoc\\
Harbin Engineering University \\
wanglin@airdoc.com}
\and
\IEEEauthorblockN{Dwarikanath Mahapatra}
\IEEEauthorblockA{IIAI \\
dmahapatra@gmail.com}
\and
\IEEEauthorblockN{Xin Zhao}
\IEEEauthorblockA{Airdoc \\
zhaoxin@airdoc.com}
\and
\IEEEauthorblockN{Mehrtash Harandi}
\IEEEauthorblockA{Monash University \\
mehrtash.harandi@monash.edu}
\and
\IEEEauthorblockN{Tom Drummond}
\IEEEauthorblockA{Monash University \\
Tom.Drummond@monash.edu}
\and
\IEEEauthorblockN{Tongliang Liu}
\IEEEauthorblockA{The University of Sydney \\
tongliang.liu@sydney.edu.au}
\and
\IEEEauthorblockN{Zongyuan Ge}
\IEEEauthorblockA{Airdoc \\
Monash University \\
zongyuan.ge@monash.edu}
}

\maketitle

\begin{abstract}
   Deep neural networks are known to be data-driven and label noise can have a marked impact on model performance. Recent studies have shown great robustness to classic image recognition even under a high noisy rate. In medical applications, learning from datasets with label noise is more challenging since medical imaging datasets tend to have asymmetric (class-dependent) noise and suffer from high observer variability. 
    In this paper, we systematically discuss and define the two common types of label noise in medical images - disagreement label noise from inconsistency expert opinions and single-target label noise from wrong diagnosis record. We then propose an uncertainty estimation-based framework to handle these two label noise amid the medical image classification task. We design a dual-uncertainty estimation approach to measure the \textbf{disagreement label noise} and \textbf{single-target label noise} via improved Direct Uncertainty Prediction and Monte-Carlo-Dropout.
    A boosting-based curriculum training procedure is later introduced for robust learning. We demonstrate the effectiveness of our method by conducting extensive experiments on three different diseases: skin lesions, prostate cancer, and retinal diseases. We also release a large re-engineered database that consists of annotations from more than ten ophthalmologists with an unbiased golden standard dataset for evaluation and benchmarking. The dataset is available at \url{https://mmai.group/peoples/julie/}.
\end{abstract}

\section{Introduction}

Deep learning is data-driven, and its success is largely attributed to sufficient human-annotated datasets. However, it is laborious to label massive data and maintain high-quality for the labels. 
In many applications, labels are acquired from non-experts (e.g., Amazon Mechanical Turk~\cite{turk2012amazon}) and sometimes automatically generated from the source information (e.g., downloading from social media with tags~\cite{guo2018curriculumnet}, extracting labels for x-ray images from associated radiology reports~\cite{wang2017chestx,irvin2019chexpert}). 
These processes could introduce potential error or label noise into the model training.  
Learning from noisy labels is a long-standing challenge and has been well-recognized in the classical image recognition problem. 
However, learning from noisy labels has not been addressed well in the medical image analysis domain~\cite{song2020learning,karimi2020deep}.
Although some previous works~\cite{jiang2018mentornet,bengio2009curriculum,li2020dividemix,shu2019meta} have had relative success to alleviate this issue, due to the domain bias and unique challenges existing in the medical imaging domain, it has limited scope for medical applications. 
For example, label noise in medical images is normally derived from the disagreement on annotations from multiple doctors, the existing methods show limited ability to handle this kind of noise. Also, most methods failed to detect noisy samples under a highly class-imbalanced scenario where \emph{noisy samples} can be easily confused with \emph{hard samples} (minority class but with clean labels) and which are eventually detected as outliers and discarded from training~\cite{xue2019robust}. 

\begin{figure}[t]
	\includegraphics[width=8cm]{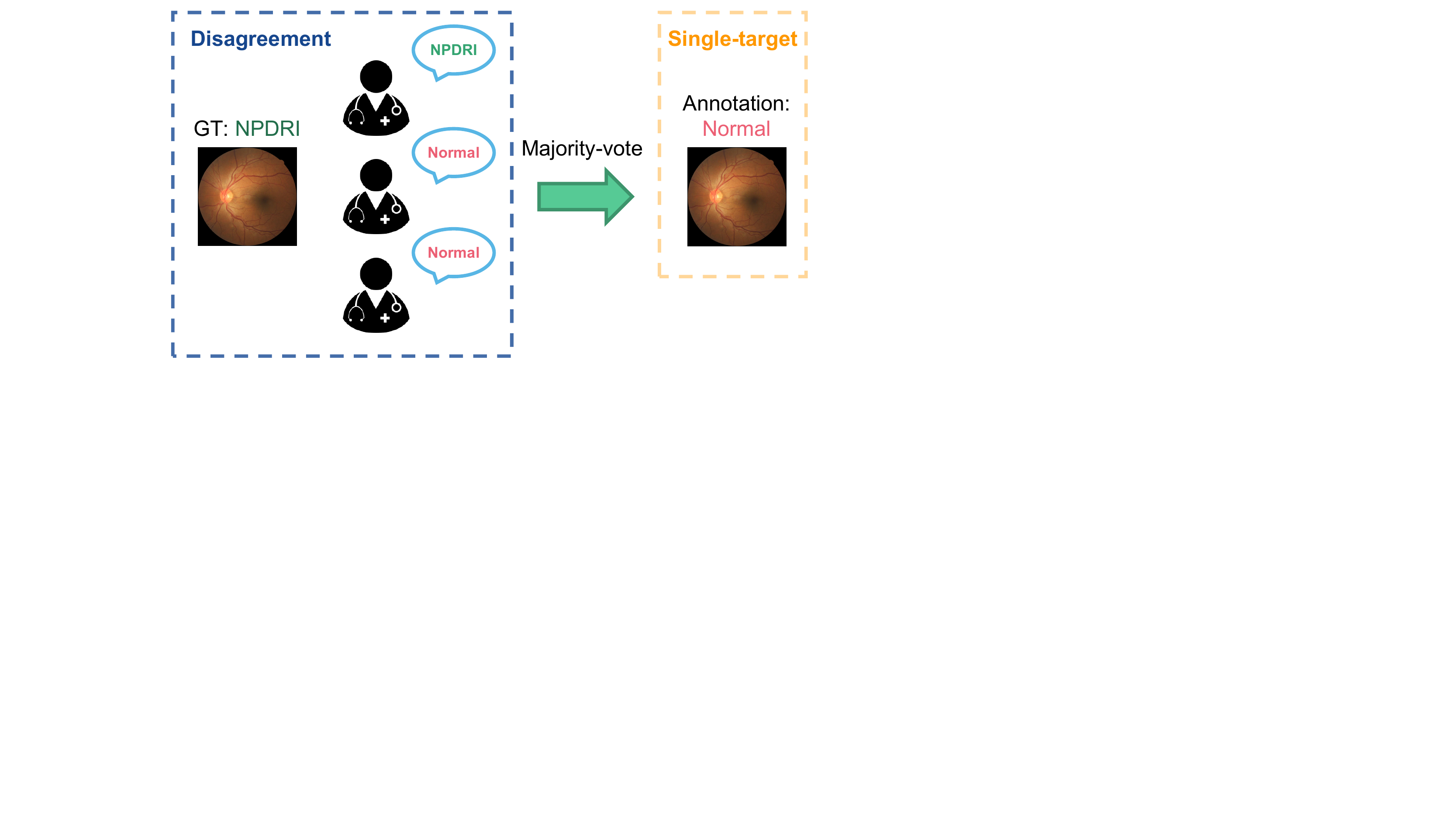}
	\centering
\caption{The illustration of two common types of label noise in medical images. For instance, the ground truth of the sample is non-proliferative diabetic retinopathy-I (NPDRI) but different opinions from multiple doctors are obtained and result in the \emph{disagreement label noise}. Then, we apply majority-vote but still get the a wrong diagnosis, which is the \emph{single-target label noise}.} \label{fig_two_noise}
\end{figure}

In this paper, we propose a dual-uncertainty-based framework for improving medical imaging classification with label noise. 
As Fig.~\ref{fig_two_noise} shows, the label noise we are trying to tackle including the following two types: (1) \emph{disagreement:} Annotations come from multiple doctors with various level of experiences, there exist disagreements and leads to observer variability; (2) \emph{single-target:} Annotations present wrong diagnostic results in nature from sole opinion.
Particularly, following the principles of reducing the impact of label noise during model training, we aim to detect those noisy samples explicitly. 
We first use the improved Direct Uncertainty Prediction (iDUP)~\cite{raghu2019direct} to capture prior knowledge to distinguish the \emph{disagreement} type noise from the noise-free label.  
Then, we use the uncertainty measurement model such as MC-Dropout~\cite{gal2016dropout} to capture the predictive uncertainty for adjudicated \emph{single-target} label using majority-vote. This will assist/help in further finding out the samples with noise on single-target labels.
Previous works~\cite{huang2019o2u,han2018co} tend to keep the low-loss/high-certainty samples retained for training and would sometimes mistakenly remove noise-free samples coming from minority class. 
In our approach, we choose to re-weight the samples according to a normalized uncertainty score, which helps preserve the impact of the minority class and the hard samples. 
Finally, a boosting-like training procedure is proposed to learn from the re-weighted samples in an intuitive and data-driven way through feeding different types of samples (clean and re-weighted samples) we have distinguished via dual-uncertainty estimation.
\begin{table}[b]
\footnotesize
\centering
\caption{The summary of some methods for some typical issues.}
\begin{tabular}{cccc}
\hline
Method  & Sample Selection      & Hard Sample & Clean Data \\ \hline
\cite{ren2018learning,jiang2018mentornet}  & Meta-Learning         & $\checkmark$         &   $\checkmark$         \\
\cite{guo2018curriculumnet}   & Subsets Clustering    & $\checkmark$         &   $\times$          \\
\cite{han2018co}  & Loss Ranking          & $\times$         &   $\times$          \\
\cite{huang2019o2u} & Learning Rate Adjustment & $ \times $ & $\times$ \\
\cite{xue2019robust} & Outlier Detection     & $\checkmark$          &  $\times$           \\
\cite{li2020dividemix}  & Gaussian Mixture Model & $\times$         &    $\times$         \\ 
Ours  & Dual-uncertainty Estimation & $\checkmark$         &    $\times$         \\ \hline
\end{tabular}
\label{table_methods}
\end{table}
Our contributions from this work are summarized as follows:
\begin{enumerate}
    \item We explore, define and provide solutions for two common types of label noise in the medical images from the perspective of \textbf{uncertainty}: \emph{disagreements} from multiple doctors and \emph{single-target} noisy labels. 
    
    \item We propose a novel dual-uncertainty-based framework which can significantly improve the robustness of the model identifying samples with label noise through measuring the uncertainty of samples. Furthermore, a boosting training procedure is proposed to handle the class-imbalance in the medical dataset. 
    
    \item We conduct extensive experiments on various imaging modalities from three kinds of diseases: skin lesions (ISIC 2019)~\cite{isic2019}, prostate cancer (Gleason 2019)~\cite{gleason2019} and retinal diseases (Kaggle DR+)~\cite{kagglediabetic}.

    \item We will release a large annotation-database containing relabelled results from more than ten doctors covering 17 retinal diseases based on Kaggle DR dataset~\cite{kagglediabetic}.  Moreover, a corresponding unbiased golden standard for evaluation and benchmark will be provided. We hope this dataset would help address the challenges of label noise in the medical image computing and computer-assisted interventions community.
\end{enumerate}

\section{Related Work}
\label{Sec. related_work}

\subsection{Deep Learning with Noisy Labels}
Deep learning with noisy labels has been an active topic in recent years and many works have been proposed to reduce the impact of label noise for practical applications~\cite{yu2017transfer,moosavi2017universal,speth2019automated}. Here, we divide the relevant methods into two main groups: sample-based and model-based. 

\noindent\textbf{Sample-based} methods aim to find out those samples whose labels are likely to be corrupted, and we can have the choice of filtering those noisy samples out or relabelling them. Various methods such as k-nearest neighbor, outlier detection, and Gaussian Mixture Model (GMM) have been widely exploited to select false-labeled samples from noisy training data~\cite{huang2019o2u, yang2020asymmetric, guo2018curriculumnet, li2020dividemix}. 
Through clustering the training data to clean samples and noisy samples, Huang~\emph{et al.}~\cite{huang2019o2u} remove the noisy samples and only train from clean data. Yang~\emph{et al.}~\cite{yang2020asymmetric} refer to Co-teaching~\cite{han2018co} and train two models to improve the robustness under a high noisy level. 
Some other works do not remove those noisy samples out: Guo~\emph{et al.}~\cite{guo2018curriculumnet} design a curriculum learning-based training procedure to help model learn the clean samples at an early stage and reduce the over-fitting to noisy data. DivideMix~\cite{li2020dividemix} transfers the noisy label challenge to a semi-supervised learning problem and leverage MixMatch~\cite{berthelot2019mixmatch} technique to more effectively learn from noisy samples. Besides, although since noisy labels are corrupted from clean labels, it has the potential to recover some useful information out of it. A typical approach for noisy labels refinement is \emph{label transition matrix estimation}. The transition matrix denotes the transition relationship from clean labels to noisy labels and the clean/noisy class posterior can be inferred using clean/noisy data~\cite{yao2020dual}. There are also some meta-learning techniques~\cite{jiang2018mentornet}, which may require some small extra clean data for reference. 

\noindent\textbf{Model-based} methods target learning to improve the robustness of the model for the noisy data. 
We review relevant works that consider network architectures, loss functions, and regularization terms. \cite{ghosh2017robust, zhang2018generalized, wang2019symmetric, lyu2019curriculum, wang2019imae} aim to design loss functions that achieve a small risk for unseen clean data even when noisy labels exist in the training data. For architectural modifications, a \emph{noise adaptation layer}~\cite{sukhbaatar2014training} is proposed to be added to the end of the network, which is equivalent to multiplication with the transition matrix between noisy and true labels. Regularization terms are usually applied to reduce the over-fitting effect and thus improve the generalization of the model. Recent works propose advanced regularization techniques such as \emph{mixup}~\cite{zhang2017mixup} and \emph{label smoothing}~\cite{pereyra2017regularizing}, that can further improve model robustness to the label noise.


\subsection{Medical Image Analysis with Noisy Labels}
\label{sec intuition}

Unfortunately, not many studies have addressed medical image classification with label noise. Here we review some works from medical imaging classification and explore some relevant potential approaches. 
Pham~\emph{et al.}~\cite{pham2019interpreting} use label smoothing technique~\cite{pereyra2017regularizing} to improve the classification of thoracic diseases from chest x-rays in the CheXpert dataset~\cite{irvin2019chexpert}. Ghesu~\emph{et al.}~\cite{ghesu2019quantifying} propose uncertainty-driven bootstrapping to filter training samples with the highest predictive uncertainty and improve robustness and accuracy on the ChestX-Ray8 dataset~\cite{wang2017chestx}. For skin lesion classification in dermoscopy images, Xue~\emph{et al.}~\cite{xue2019robust} proposed an online uncertainty sample mining method and a sample re-weighting strategy to preserve the usefulness of correctly-labeled hard samples. Dgani~\emph{et al.}~\cite{dgani2018training} leverage the noise adaptation layer~\cite{goldberger2016training} on mammography classification task and outperform standard training methods.



Although some approaches have been considered to address the noisy label issue, we still observe a noticeable gap in applying those methods to medical image classification with label noise. Some of those methods will serve as baselines in this work. 
Meta-learning-based approaches~\cite{ren2018learning,jiang2018mentornet} require a small clean validation dataset on the side to adjust the weights for samples in a mini-batch manner. 
CurriculumNet~\cite{guo2018curriculumnet} leverages density-distance~\cite{rodriguez2014clustering} and KMeans~\cite{krishna1999genetic} to cluster samples to different subsets according to the complexity, then perform training from the subsets respectively. However, the functional ability of CurriculumNet on a small dataset with high-similarity has not been evaluated yet. 
Co-teaching~\cite{han2018co} builds two CNN models, and the samples with the smallest forward-propagation loss in a mini-batch are fed into the other network for the next round of training. However, in the medical dataset, the amount of dataset of each class is usually imbalanced, and those high loss samples are always from minority class and treated as hard samples. Removing those samples would only make the situation worse. 
Xue~\emph{et al.}~\cite{xue2019robust} do not only remove the noisy samples but also re-weight all samples using a probabilistic Local Outlier Factor algorithm
(pLOF)~\cite{kriegel2009loop} to retain the hard samples. 
DivideMix~\cite{li2020dividemix} is proposed to find out potentially noisy samples through GMM and then transfer it into a semi-supervised problem. However, there still lacks direct insights for handling imbalanced data, and the calculated-weights is not available for the ``real" unlabeled data. 
In Table~\ref{table_methods}, we give a summary of those methods. \textbf{Sample Selection} denotes the auxiliary technique being used for sample selection; \textbf{Hard Sample} denotes whether the method is capable of handling the hard samples appropriately during the sample selection process; \textbf{Clean Data} indicates whether the method requires extra clean data for training or model calibration.


\section{Methodology}
\label{Sec. methods}
\subsection{Problem Definition}
Given the $i_{th}$ sample $x_{i}$ in the dataset $X = [x_{1}, x_{2}, ..., x_{z}]$ from the real clinical scenario, we have pairs of the form (sample, annotations from multiple doctors), ($x_{i}; y^{1}_{i}, y^{2}_{i}, ..., y^{n_{i}}_{i}$), where $n_{i}$ denotes the number of annotations received for the current sample. For example, $n_{i} = 1$ means the sample is labeled by only one doctor while $n_{i} > 1$ indicates the sample is labeled by multiple doctors. Label noises may occur due to the inconsistency opinions from doctors or simply wrong annotation.  

\begin{figure*}[t]
	\includegraphics[width=17cm]{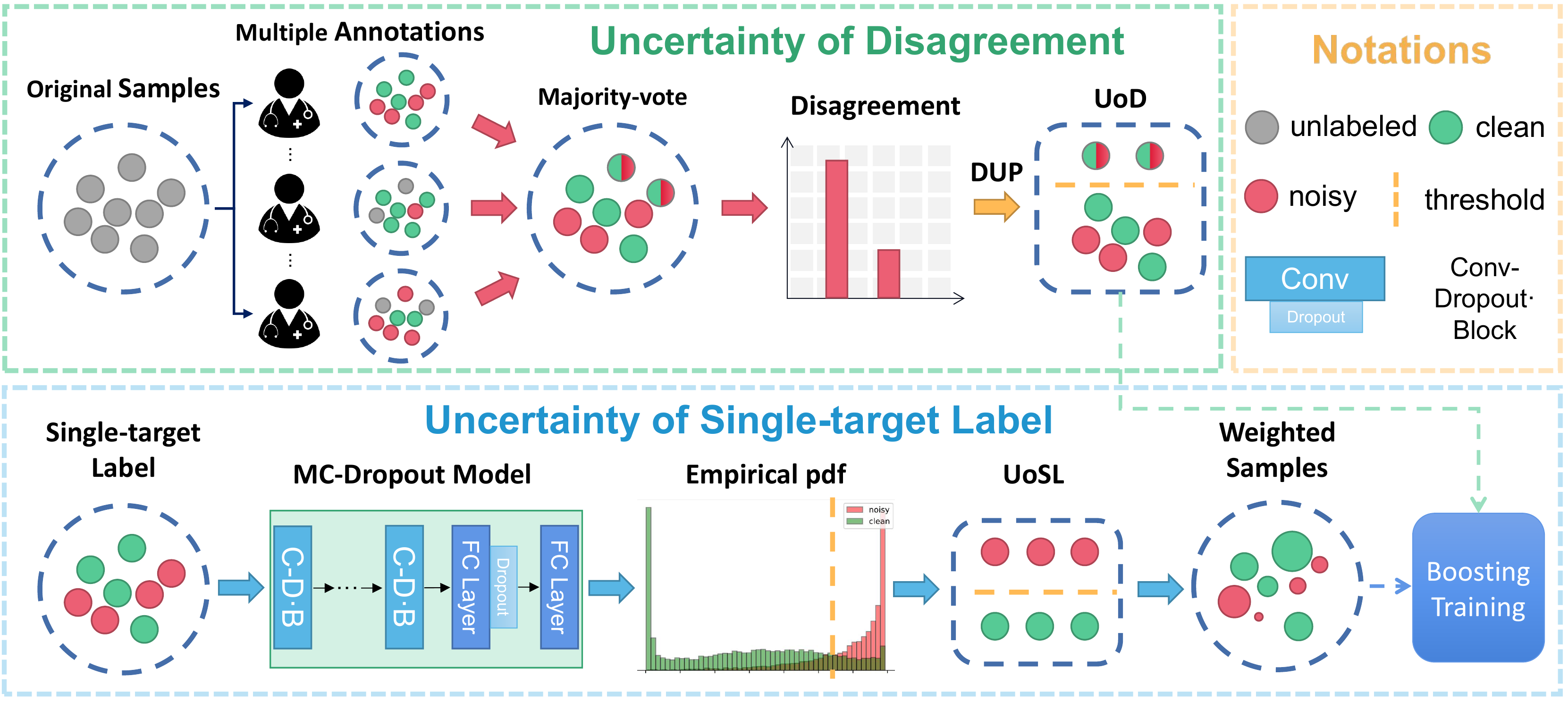}
	\centering
\caption{Here shows the overview of our proposed dual-uncertainty estimation framework. For those samples with annotations from multiple doctors, we use improved Direct Uncertainty Prediction to estimate the \textbf{UoD} (uncertainty of disagreement), and a threshold is set to filter out those samples with high \textbf{UoD}. Then given a sample with an adjudicated single-target label, we leverage the Monte Carlo estimate MC-Dropout model to generate predictive uncertainty scores to estimate the \textbf{UoSL} (uncertainty of single-target label) and re-weight those potential noisy samples. Those selected samples will then fed into a boosting training scheme to train a noisy label tolerant classifier. } \label{fig_framework}
\end{figure*}

Hence, under the circumstances where samples with label noise presented, our goal in this work is to train a classification model $h$ with parameters $\theta$ from modified training samples $\hat{X} = [\hat{x}_{1}, \hat{x}_{2}, ..., \hat{x}_{\hat{z}}]$ based on our proposed selection/weighting criteria and evaluate its performance on an unbiased golden standard 
test set. We give an overview of our proposed framework in Fig.~\ref{fig_framework}

\subsection{Dual-uncertainty Estimation}

We carry out two kinds of uncertainty as Fig.~\ref{fig_two_noise} shows. We define the first one as Uncertainty of Disagreement (UoD) 
, which signifies expert disagreement among multiple doctors. And the second one is defined as Uncertainty of Single-target Label to signify the noise of single adjudicated annotation. We propose to use dual-uncertainty estimation for modeling this two uncertainty respectively.

\subsubsection{Uncertainty of Disagreement}
\label{UoD}
By applying aggregation methods such as average voting on a sample with high disagreement among annotators, it tends to generate noisy training labels. 
However, we may still extract useful information or clues out of these annotations from those disagreements to detect hard samples vs wrong labels since those samples with high disagreement are considered to be easily misdiagnosed by human-experts, also difficult to be learned by DL models.  
Suppose we can find an uncertainty estimation approach $U_{1}$ to map and quantify the annotations $Y_{i} = [y^{1}_{i}, ..., y^{n_{i}}_{i}]$ for the sample $x_{i}$ to the UoD, and we can handle those samples with an uncertainty metric and then can be served as a kind of prior knowledge.

To provide a quantified score for the uncertainty of disagreement, Direct Uncertainty Prediction (iDUP)~\cite{raghu2019direct} is considered to be an unbiased estimate of the true uncertainty which is based on the theory of \emph{Modeling Labeler Bias}~\cite{wauthier2011bayesian}. Letting $G = [g_{1}, g_{2}, ..., g_{k}]$ denotes the different diseases/grades (e. g,. $k=5$ represents there are five levels of severity for DR~\cite{kagglediabetic}), and $P_{i} = [p^{1}_{i}, p^{2}_{i}, ..., p^{k}_{i}]$ represents the empirical grade distribution, also known as the empirical histogram for the sample $x_{i}$ is calculated as:
\begin{equation}
    p^{c}_{i} = \frac{\sum _{j}\mathbbm{1_{y^{j}_{i} = g^{c}}}}{n_{i}}
\end{equation}
By applying DUP, the original UoD is computed as:
\begin{equation}
    UoD_{i} = U_{1}(x_{i}) = U(P_{i}) = 1 - \sum_{j=1}^{k} (p^{j}_{i})^{2},
\end{equation}

Although DUP is proven to be effective to handle the UoD in a dataset with annotations from multiple doctors, it has obvious limitations. For example, given two samples $x_{1}$ and $x_{2}$ labeled with $Y_{1} = [0,1]$ and $Y_{2} = [0,0,0,1,1,1]$ from 2 and 6 doctors respectively, we can calculate the empirical histogram as $P_{1} = [0.5,0.5]$ and $P_{2} = [0.5,0.5]$. For two samples, they obtain the same $UoD$ score: $1-0.5^{2}-0.5^{2} = 0.5$. However, there is a possibility that the latter should be more uncertain than the former. If extra doctors are introduced, the results of $UoD$ may change, and DUP do not have the ability to handle this potential uncertainty.

Here, we propose to take the variance of the number of annotations into consideration by adding a factor to $UoD$:
\begin{equation}
    iUoD_{i} = [min(\sum _{j}\mathbbm{1_{y^{j}_{i} = g^{c}}})]^{\eta} \cdot UoD_{i},
\end{equation}
where $\eta$ is a hyper-parameter. We use the value of the category with the least number of votes as a factor.

During training, we can set a threshold $t_{1}$ to select a fraction of training samples with relatively high uncertainty ($UoD > t_{1}$) to be eliminated. 
With pre-calculated UoD as the prior knowledge, we can know which samples are likely to be misdiagnosed by human-experts, also difficult to be learned or fitted by deep learning models. 

\subsubsection{Uncertainty of Single-target Label}
Estimation of UoD learns an uncertain score from the raw data pairs 
and has provided a kind of prior knowledge 
for sample selection 
However, we have not tackled the single-target label noise for those samples where the diagnosis from single or the majority-vote result still can be wrong.
Previous works~\cite{guo2018curriculumnet,han2018co,xue2019robust,ren2018learning,huang2019o2u,li2020dividemix} proposed several techniques to reduce the impact of wrongly-labeled samples in trainset by sample selection.
However, those methods lack indirect insights on the imbalanced attribute existed in the medical imaging domain, as we discussed in Sec.~\ref{sec intuition}. 
To put data imbalance attribute into consideration while addressing the single-target noisy label issue, we propose to use uncertainty estimation techniques (e, g., Monte-Carlo Dropout~\cite{gal2016dropout}). 
More specifically, we perform $T$-times
stochastic forward pass on a trained CNN model under random dropout, then we have a pseudo-decision distribution from $T$ "CNN doctors" with various degrees of disagreement. 
Based on this pseudo-decision distribution by several "CNN doctors", we can leverage the uncertainty of disagreement discussed in Sec.~\ref{UoD} to detect potential outliers that have a bad impact on the model's performance.

Formally, for T disease probability vectors for a subject $x_{i}: \{p^{(t)}_{i}\}^{T}_{1}$. The uncertainty of a single-target label can be estimated using the mean predictive entropy~\cite{kendall2017uncertainties}: 
\begin{equation}
    UoSL_{i} = -\sum _{c}m_{i,c}log m_{i,c}
\end{equation}
where $m_{i} = \frac{1}{T}\sum _{t}p^{t}_{i}$ and $c$ corresponds to the $c$-th class. Mean predictive entropy shows a stronger ability to make representations of variance from MC estimators compared to softmax distributions. Please check~\cite{kendall2017uncertainties} for more details.

\subsection{Boosting Training}

\subsubsection{Sample selection} Given dual-uncertainty score $UoD_{i}$ and $UoSL_{i}$ for each sample $x_{i}$, we propose to use these uncertainty measurements for sample selection and re-weighting~\cite{freund1999short} at each training iteration. 
Firstly, for those samples with multiple annotations, we obtain $UoD_{i}$. Specially, $UoD_{i} = 1$ for those sample with single-target label from only one doctor.
A threshold $t_{UoD}$ is set to filter out samples with a high score of $UoD_{i}$.
Those samples generally show significant controversies from the annotators and are diminishing for the model training. 
Similarly, we assign scalar weights to dynamically adjust the samples with adjudicated single-target labels whose predictive uncertainty score are above the threshold $t_{UoSL}$. 
We then reorder the samples for each sample $x_{i}$ in $X$ 
according to the results of UoD estimation as follows: $UoD{x_{1}} \leq UoD{x_{2}} \leq ... \leq UoD{x_{z}}$. 
Finally, we have the retained $\hat{n}_{i}$ out of $n_{i}$ samples $\hat{X} = [(\hat{x}_{1};\hat{y}_{1}), (\hat{x}_{2};\hat{y}_{2}), ..., (\hat{x}_{\hat{n}_{i}};\hat{y}_{\hat{z}})]$.  
However, it is notated that the elimination of hard samples from the minority classes is still unavoidable since they are likely to receive a high $U$ score. Then, a curriculum training strategy is proposed to reduce the impact of removing hard samples.


\subsubsection{Curriculum training} We feed the selected clean samples $\hat{X}$ to the model at an early stage 
and then we have the network to begin learning all samples with a normalized weights 
in a curriculum-driven way~\cite{bengio2009curriculum}. Overall, the network is jointly optimized by the following loss functions:
\begin{equation}
\label{eq fl}
    L_{FL}(\hat{X}) = -\sum^{\hat{z}}_{i=1}(1-p_{i}\cdot q_{i})^{\gamma}\cdot log(p_{i}),
\end{equation}
\begin{equation}
    L_{wCE}(X) = -\sum^{z}_{i=1} w_{i} \cdot (q_{i}\cdot \log p_{i} + (1-q_{i})\cdot \log (1-p_{i}))
\end{equation}
\begin{equation}
   L = \alpha \times L_{FL}(\hat{X}) + \beta  \times L_{wCE}(X),
\end{equation} 
where $p_{i}$ denotes the predictions and $q_{i}$ denotes the ground-truths. $\alpha$ and $\beta$ are hyper-parameters as loss weights. We leverage the Focal Loss~\cite{lin2017focal} in Eq.~\ref{eq fl} for reducing the impact of imbalance, and assign the weights driven from the uncertainty measurement $W = [w_{1},...,w_{i},...,w_{z}]$ to the Cross-Entropy Loss for all samples 
, where $nUoSL$ is the normalized $UoSL$ mapped to [0, 1]:
\begin{equation}
w_{i} = \begin{cases}
1 - nUoSL_{i}, & nUoSL_{i} > t_{UoSL} \\
1, & nUoSL_{i} \leq t_{UoSL}\\
\end{cases}
\end{equation}

\section{Experiments}
\label{Sec. Experiments}
In this section, to demonstrate the complex scenarios in the clinical collection of datasets, we conduct the experiments on three medical imaging datasets, including (1) ISIC 2019 dataset~\cite{isic2019} for skin cancer diagnosis, (2) Gleason 2019~\cite{gleason2019} for prostate tissue microarray classification, (3) Kaggle+ fundus images database released by this paper, consists of 17 diseases and annotations from multiple ophthalmologists. 
In the experiments, we explore \emph{single-target} noisy diagnosis scenario for the ISIC 2019 dataset and \emph{disagreement/observer-variability} scenario for the Gleason 2019 and Kaggle+ fundus. 


\subsection{Implementation Details}
Following~\cite{han2018co, huang2019o2u}, we use Dropout-ResNet-101\footnote{a dropout layer with the probability of 0.3 is inserted after every \textbf{Basic Layer}}~\cite{he2016deep} and a 9-Layer CNN with dropout layer \footnote{a dropout layer with the probability of 0.3 is inserted after every convolution layer and the first fully connected layer} as backbones for a fair comparison. The batch sizes are set to 16. 
We only use horizontal flipping as the data augmentation strategy since the overuse of various data augmentation techniques may have a negative impact on the uncertainty measurement~\cite{ayhan2020expert}. 
We apply the Adam optimizer for model training, and the initial learning rate is set to 3e-4 and decayed by a factor of 0.5 when there are no more improvements in the performance with the patience of 5. $\alpha$ is set as 1 and $\beta = (\frac{epoch_{i}}{epoch_{all}})^{2}$ with the curriculum training progress. $epoch_{all}$ is set as 5, 5 and 8 for ISIC 2019, Gleason 2019 and Kaggle DR+ respectively. $t_{UoD}$ is set as 0.5. Experiments were run on 8 $\times $ NVIDIA GTX 1080Ti graphic cards.

\subsection{ISIC 2019}
\subsubsection{Dataset Statistics and Noise Synthesizing}
The ISIC 2019 dataset contains dermoscopic images for skin cancer diagnosis across 8 different categories and
each sample also gets a binary diagnosis result of benign or malignant. 
In this study, 25,331 images (16,971 benign and 8,360 malignant) are available for training.
Regarding the synthesizing method, we followed ~\cite{xue2019robust} to build the asymmetric label noise for the ISIC 2019 dataset based on the training loss\footnote{See our Appendix for more details}.



\newcommand{\mysize}{3.7cm}
\begin{figure*}[t]
\centering

\subfigure[Confidence Penalty Epoch 5]{
\includegraphics[width=4.0cm]{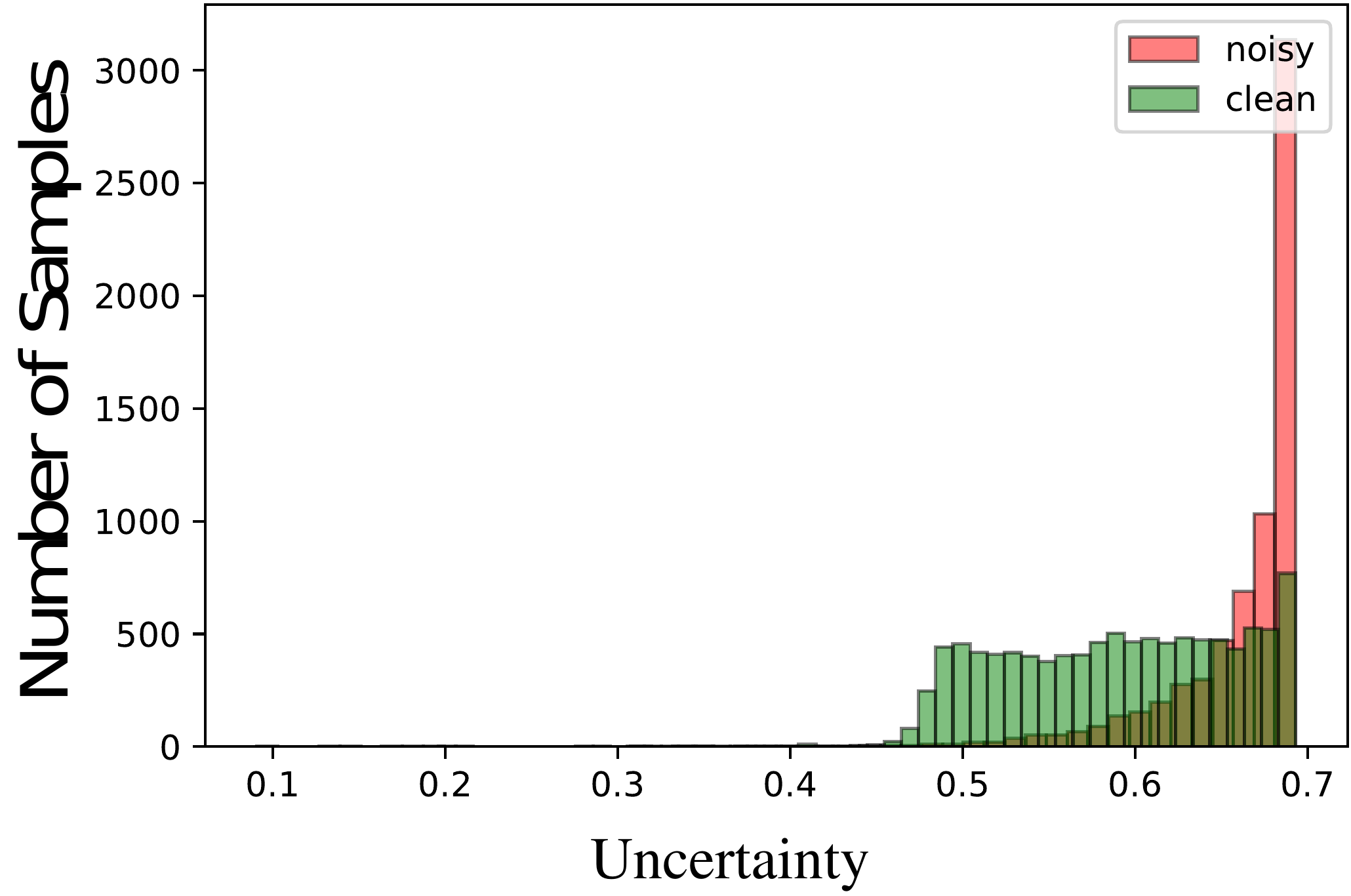}
}
\quad
\subfigure[MC-Dropout Epoch 1]{
\includegraphics[width=\mysize]{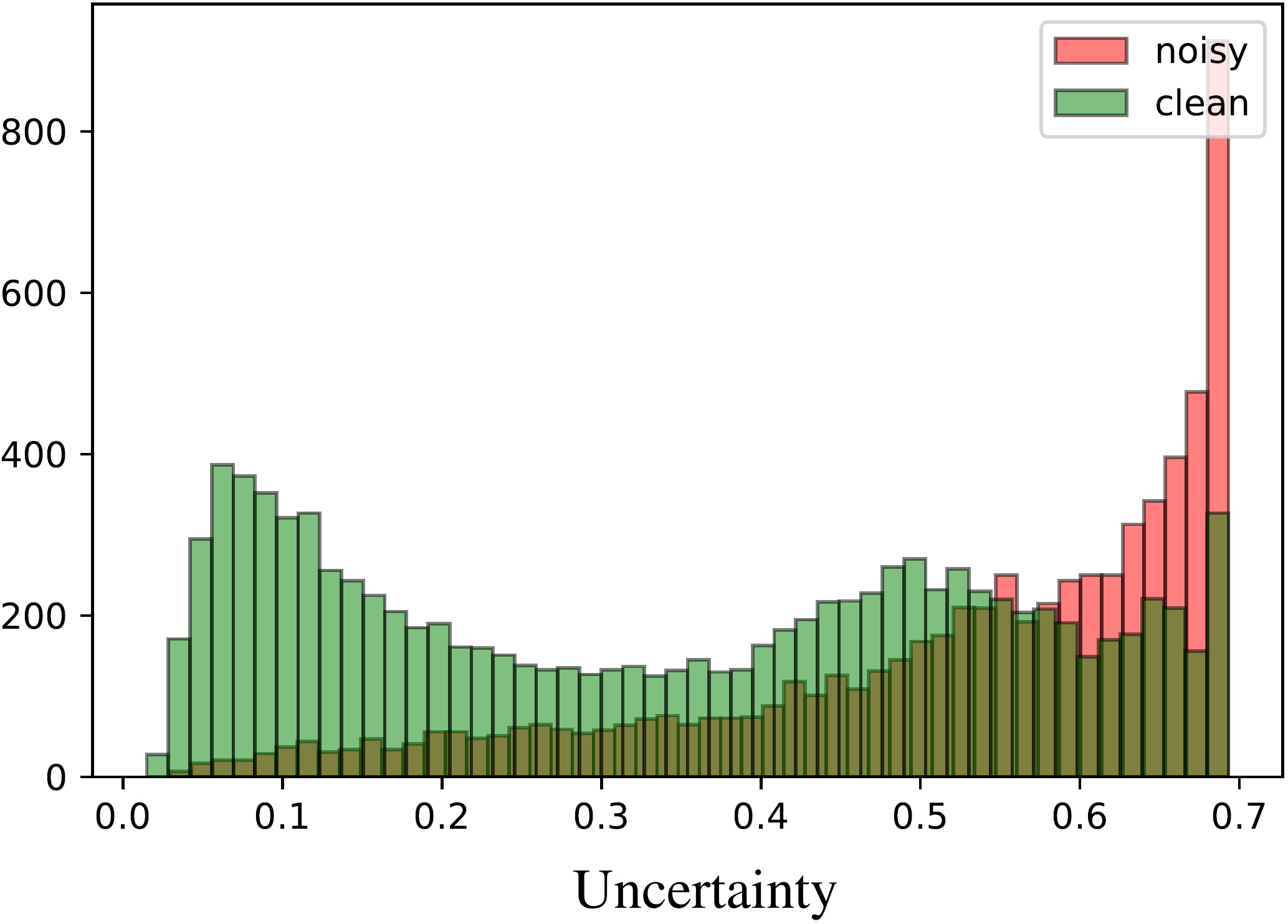}
}
\quad
\subfigure[MC-Dropout Epoch 5]{
\includegraphics[width=\mysize]{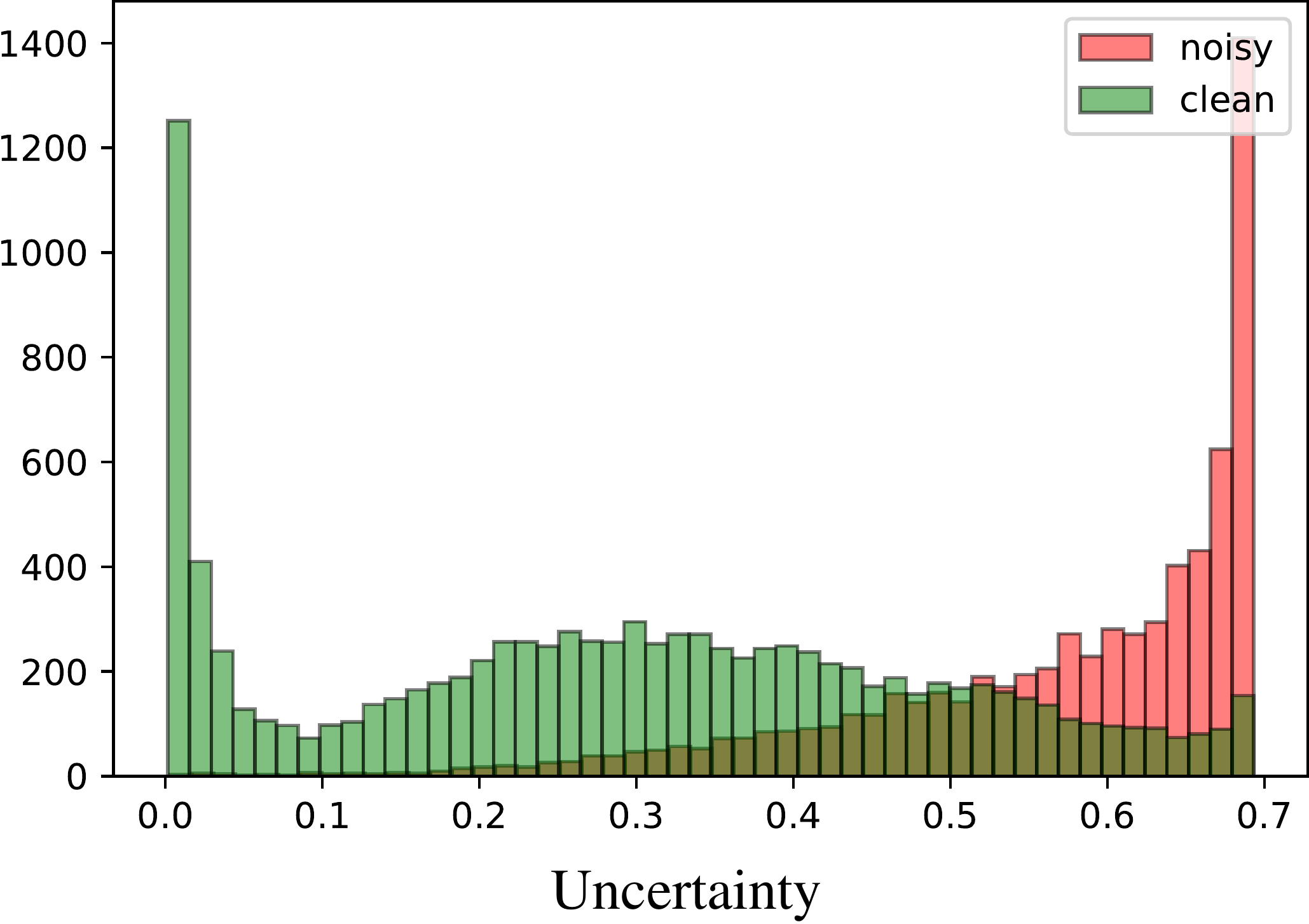}
}
\quad
\subfigure[MC-Dropout Epoch 15]{
\includegraphics[width=\mysize]{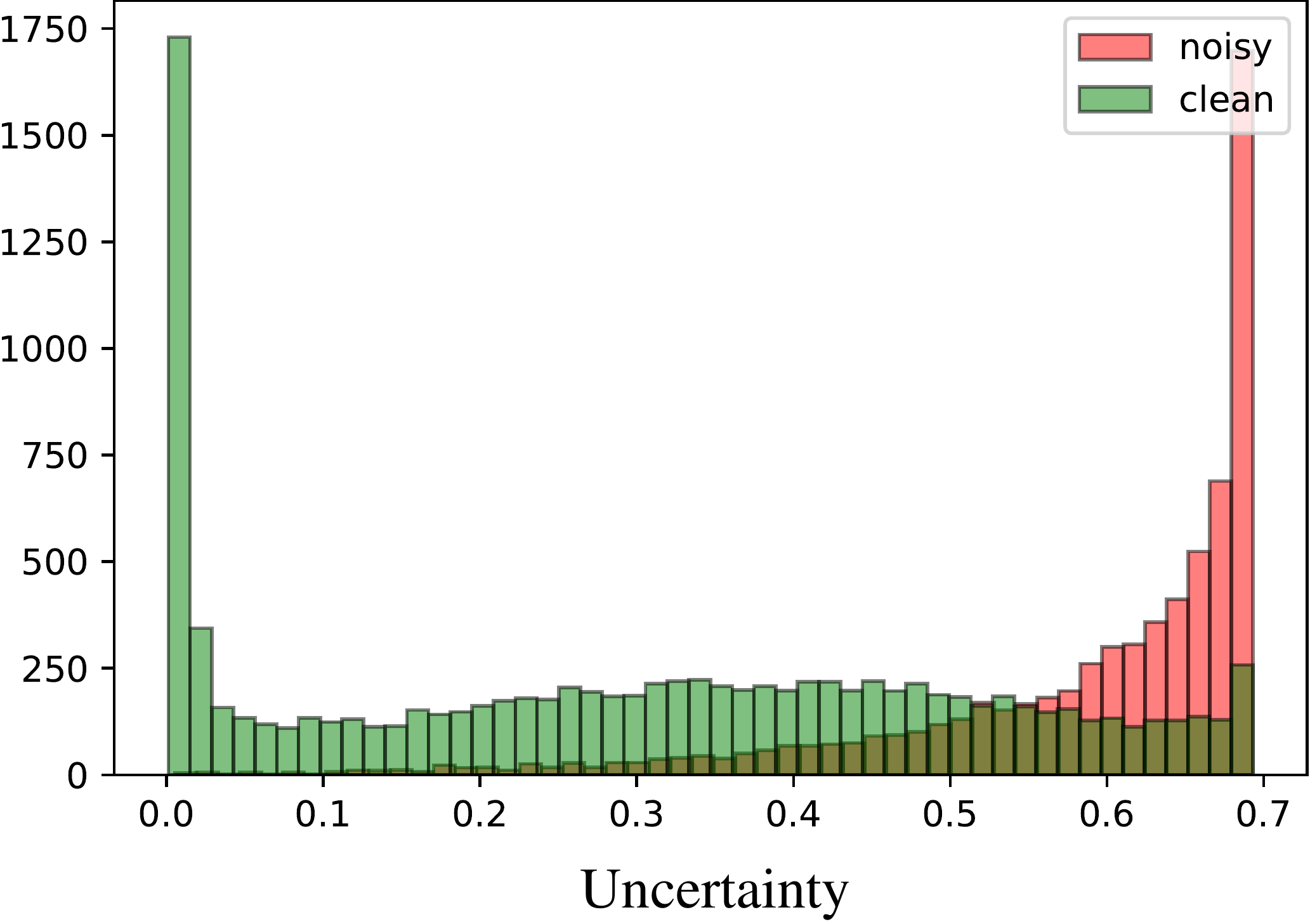}
}

\caption{The visualization results on ISIC 2019. (a) Warmup using Confidence Penalty. (b) - (d) Warmup using MC-Dropout.} 
\label{fig. uncertainty}
\end{figure*}
\subsubsection{Warmup} For the initial estimation of the predictive uncertainty score, the warmup procedure is needed for a few epochs by training on all data points using the normal cross-entropy loss. Unlike classic imaging datasets, label noise in medical images is asymmetric (i.e., class-conditional) due to the high similarity between categories. The model could be quickly overfitting to the wrong information at an early stage and give low-entropy predictions. 
DivideMix~\cite{li2020dividemix} proposed to reduce the speed of overfitting to incorrect labels by adding a negative entropy term\cite{pereyra2017regularizing} to the loss function. In Fig.~\ref{fig. uncertainty}, we first conduct experiments on two warmup strategies and show the comparative results between confidence penalty-based and uncertainty-based technique, as this will reveal models' ability for noisy samples detection. We can see that although the confidence penalty prevents the model from generating low-entropy predictions, it makes the model difficult to learn the correct information from clean samples (see Fig.~\ref{fig. uncertainty} - (a)) and a good trade-off strategy is needed between noisy samples and clean samples after assessing the uncertainty score.

 

\subsubsection{Comparative Study} We compare our proposed method with baselines~\cite{zhang2017mixup,pereyra2017regularizing,guo2018curriculumnet,han2018co,huang2019o2u} discussed in Sec.~\ref{sec intuition} and re-implemented using the same network backbone architecture. Although those methods are designed and evaluated for classic image recognition tasks, here we re-adapted them under well-adjusted hyper-parameters on our target medical dataset for a fair comparison.  It is widely ac-cepted that the model tends to learn clean samples in the early stage and gradually overfit to noisy samples,  which results in a performance drop. Thus, we follow~\cite{li2020dividemix} and report the best percentage AUC score across all the epochs and average results over the last three epochs, to evaluate the robustness of our proposed framework, which are denoted by \textbf{B} and \textbf{L} respectively in Table~\ref{table isic results}. 
We also give the differential score between the best epoch and the last epoch. 
The results show that our proposed framework exceeds all other baselines (in \textbf{bold}) with both the best and last epoch under all various noisy ratios except 10\%. 
The reason is that training on Cross-Entropy and MixUp methods do not change the original distribution of samples significantly, so they can achieve more stable results under the low noisy rate, which is equivalent to be trained on a clean dataset. 
The performance difference between the best epoch and the last epoch demonstrates the robustness of each method. Without losing the performance of direct training, our proposed framework achieves 1.90\% and 1.03\% improvement compared to CurriculumNet under the noisy rate of 10\% and 20\%, 0.28\% compared to MixUp, 0.68\% compared to Label Smoothing and 4.02\% compared to CurriculumNet under the 40\% label noise. 


\begin{table}[t]
\scriptsize
\renewcommand\arraystretch{1.2}
\centering
\caption{The percentage AUC results on ISIC 2019.}
\begin{tabular}{p{1.6cm}ccccc}
\hline
\multicolumn{6}{c}{ISIC 2019 (Benign / Malignant)}                                                                             \\ \hline
         & \multicolumn{1}{c|}{}     & \multicolumn{3}{c}{ResNet-101}                   & 9-layers CNN \\ \hline
     & \multicolumn{1}{c|}{}     & 10\%    & 20\%    & \multicolumn{1}{c|}{40\%}    & 40\%         \\ \hline
\multirow{2}{*}{Cross-Entropy}        & \multicolumn{1}{c|}{B} & $82.63_{\Longleftrightarrow}$ & $80.73_{\Longleftrightarrow}$ & \multicolumn{1}{c|}{$78.27_{\Longleftrightarrow}$} & $76.12_{\Longleftrightarrow}$      \\
                                 & \multicolumn{1}{c|}{L} & $80.69^{1.94}$ & $78.71^{2.02}$ & \multicolumn{1}{c|}{$76.66^{1.61}$} & $74.23^{1.89}$      \\ \hline
\multirow{2}{*}{MixUp~\cite{zhang2017mixup}}           & \multicolumn{1}{c|}{B} & $81.99_{\Longleftrightarrow}$ & $80.83_{\Longleftrightarrow}$ & \multicolumn{1}{c|}{$79.49_{\Longleftrightarrow}$} & $76.17_{\Longleftrightarrow}$      \\
                                 & \multicolumn{1}{c|}{L} & $\textbf{81.59}^{0.40}$ & $ 77.85^{2.98}$  & \multicolumn{1}{c|}{$76.86^{2.63}$} & $74.45^{1.72}$      \\ \hline
\multirow{2}{*}{LS~\cite{pereyra2017regularizing}} & \multicolumn{1}{c|}{B} & $81.24_{\Longleftrightarrow}$ & $80.56_{\Longleftrightarrow}$ & \multicolumn{1}{c|}{$ 79.65_{\Longleftrightarrow}$ } & $77.00_{\Longleftrightarrow}$      \\
                                 & \multicolumn{1}{c|}{L} & $79.35^{1.89}$ & $78.30^{2.26}$ & \multicolumn{1}{c|}{$76.62^{3.03}$} & $74.78^{2.22}$      \\ \hline
\multirow{2}{*}{CurriculumNet~\cite{guo2018curriculumnet}}   & \multicolumn{1}{c|}{B} & $82.65_{\Longleftrightarrow}$ & $81.12_{\Longleftrightarrow}$ & \multicolumn{1}{c|}{$79.38_{\Longleftrightarrow}$} & $77.05_{\Longleftrightarrow}$      \\
                                 & \multicolumn{1}{c|}{L} & $79.12^{4.53}$ & $76.23^{4.89}$ & \multicolumn{1}{c|}{$73.01^{6.37}$} & $71.09^{5.96}$      \\ \hline
\multirow{2}{*}{Co-teaching~\cite{han2018co}}     & \multicolumn{1}{c|}{B} & $80.25_{\Longleftrightarrow}$ & $78.80_{\Longleftrightarrow}$ & \multicolumn{1}{c|}{$77.23_{\Longleftrightarrow}$} & $75.02_{\Longleftrightarrow}$      \\
                                 & \multicolumn{1}{c|}{L} & $79.64^{0.61}$ & $77.61^{1.19}$ & \multicolumn{1}{c|}{$75.11^{2.12}$} & $73.14^{1.88}$      \\ \hline
\multirow{2}{*}{O2U-Net~\cite{huang2019o2u}}         & \multicolumn{1}{c|}{B} & $77.14_{\Longleftrightarrow}$ & $76.37_{\Longleftrightarrow}$ & \multicolumn{1}{c|}{$74.04_{\Longleftrightarrow}$} & $72.14_{\Longleftrightarrow}$      \\
                                 & \multicolumn{1}{c|}{L} & $74.55^{2.59}$ & $74.12^{2.25}$ & \multicolumn{1}{c|}{$73.97^{0.07}$} & $71.56^{0.58}$      \\ \hline
\multirow{2}{*}{Ours}            & \multicolumn{1}{c|}{B} &  $\textbf{83.26}_{\Longleftrightarrow}$       &   $\textbf{82.90}_{\Longleftrightarrow}$    & \multicolumn{1}{c|}{$\textbf{81.01}_{\Longleftrightarrow}$}      &  $\textbf{{80.34}}_{\Longleftrightarrow}$          \\
                                 & \multicolumn{1}{c|}{L} & $80.63^{2.63}$      &   $\textbf{79.04}^{3.86}$   & \multicolumn{1}{c|}{$\textbf{ 78.66}^{2.35}$}      &  $\textbf{{76.21}}^{4.13}$           \\ \hline
\end{tabular}

\label{table isic results}
\end{table}


\subsection{Gleason 2019}
\subsubsection{Dataset Statistics}
Gleason 2019 aims to classify prostate tissue microarray (TMA) cores as one of the four classes: benign and cancerous with Gleason grades 3, 4, and 5. TMA cores have been annotated in detail (i.e., pixel-wise) by six pathologists independently. Because pathologists who labeled this dataset have a different level of experience, ranging from 1 to 27 years, \emph{uncertainty of disagreement} presents among the samples. It is also noted that not all images are labeled by the exact same doctors (i.e., doctor ID 6 only labeled 65 images out of all 244 images).
To augment the number of available training and testing samples from the Gleason 2019 dataset, 
as suggested by~\cite{arvaniti2018automated}, a patch-level classification system are normally built to verify the idea.  
For patch generation, each whole-slide image is first resized to 3100 $\times$ 3100 pixels, then small image regions of size 750 x 750 are sampled from each TMA spot, using a step-size of 375 pixels\footnote{See more details of building datasets in our Appendix}. 
We give the processed data statistics in Table~\ref{table_gleason}.
We use two metrics: quadratic weighted kappa~\cite{fleiss1973equivalence} and Fleiss' kappa~\cite{fleiss1969large} to quantify the observer variability between each pathologist and majority-vote results in Table~\ref{Table_kappa_gleason}.



\begin{table}[t]
\footnotesize
\centering
\renewcommand\arraystretch{1.2}
\caption{The Data statistics on Gleason 2019. \textbf{Origin} are counted in the image-level and others are in patch-level.}
\begin{tabular}{cp{0.5cm}<{\centering}p{0.5cm}<{\centering}p{0.7cm}<{\centering}p{0.7cm}<{\centering}p{0.7cm}<{\centering}p{0.7cm}<{\centering}p{0.7cm}<{\centering}}
\hline
Target        & Origin & Benign & Gleason 3 & Gleason 4 & Gleason 5 & Sum   \\ \hline
Pathologist 1 & 244    & 210    & 1513      & 4166      & 617       & 6506  \\
Pathologist 2 & 141    & 614    & 612       & 827       & 369       & 2422  \\
Pathologist 3 & 242    & 776    & 2165      & 3190      & 14        & 6135  \\
Pathologist 4 & 244    & 1521   & 2183      & 3430      & 66        & 7200  \\
Pathologist 5 & 246    & 940    & 2946      & 3411      & 99        & 7396  \\
Pathologist 6 & 65     & 409    & 652       & 826       & 174       & 2060  \\
Majority-vote         & -      & 1835   & 3202      & 5342      & 255       & 10634 \\
Test         & -      & 470   & 781      & 1161      & 27       & 2439 \\ \hline
\end{tabular}
\label{table_gleason}
\end{table}




\begin{figure}[t]
\centering
	\includegraphics[width=8cm]{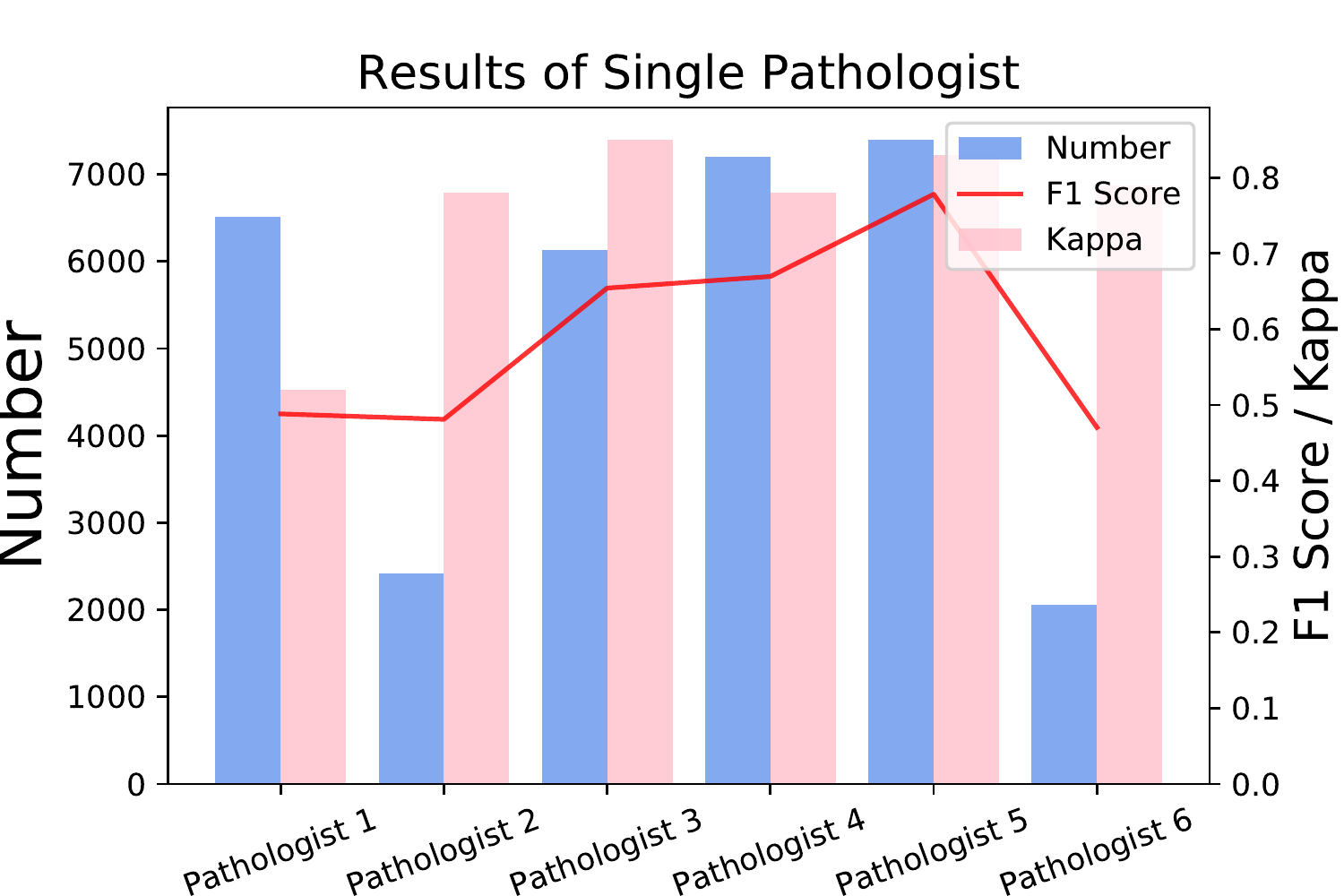}
\caption{The performance (F1 Score in red line) differs significantly with respect to pathologists' annotations (Kappa, in pink bars) and the number of training data (in blue bars).} \label{fig_sp}
\end{figure}

\begin{table}[t]
\scriptsize
\centering
\renewcommand\arraystretch{1.2}
\caption{Cohen's and Fleiss' kappa value of Gleason 2019.}
\begin{tabular}{cp{0.3cm}<{\centering}p{0.3cm}<{\centering}p{0.3cm}<{\centering}p{0.3cm}<{\centering}p{0.3cm}<{\centering}p{0.3cm}<{\centering}p{1.6cm}<{\centering}}
\hline
Name  & 1   & 2   & 3   & 4   &  5   &  6   & Fleiss' Kappa \\ \hline
Value & 0.52 & 0.78 & 0.85 & 0.78 & 0.83 & 0.79 & 0.33          \\ \hline
\end{tabular}
\label{Table_kappa_gleason}
\end{table}

\begin{table}[t]
\scriptsize
\centering
\caption{The results of Gleason 2019.}
\begin{tabular}{lcccc}
\hline
\multicolumn{5}{c}{Gleason 2019}                                                                 \\ \hline
\multicolumn{1}{c}{}                & \multicolumn{1}{c|}{}     & \multicolumn{3}{c}{ResNet-101} \\ \hline
\multicolumn{1}{c}{}                & \multicolumn{1}{c|}{}     & Recall  & Precision & F1 Score \\ \hline
\multirow{2}{*}{SP}                 & \multicolumn{1}{c|}{Best} & 63.53   & 68.10     & 59.03    \\
                                    & \multicolumn{1}{c|}{Last} & 61.72   & 65.37     & 57.74    \\ \hline
\multirow{2}{*}{SP + Ensemble}      & \multicolumn{1}{c|}{Best} & 69.36   & 84.17     & 72.46    \\
                                    & \multicolumn{1}{c|}{Last} & 67.01   & 70.32     & 65.01    \\ \hline
\multirow{2}{*}{SP + PL}            & \multicolumn{1}{c|}{Best} & 75.12   & 70.13     & 70.31    \\
                                    & \multicolumn{1}{c|}{Last} & 72.33   & 70.34     & 69.15    \\ \hline
\multirow{2}{*}{SP + PL + Ensemble} & \multicolumn{1}{c|}{Best} & 80.02   & \textbf{85.43}     & 82.40    \\
                                    & \multicolumn{1}{c|}{Last} & 76.71   & \textbf{80.90}     & 78.48    \\ \hline
\multirow{2}{*}{SP + SO + Ensemble}& \multicolumn{1}{c|}{Best} & \textbf{82.43}   & 81.33     & 81.88    \\
                                    & \multicolumn{1}{c|}{Last} & \textbf{80.96}   & 80.80     & \textbf{80.87}    \\ \hline
\multirow{2}{*}{Annotator confusion estimation}      & \multicolumn{1}{c|}{Best} & 80.31   & 81.86     & 81.08    \\
                                    & \multicolumn{1}{c|}{Last} & 80.21   & 80.98     & 80.59    \\ \hline
\multirow{2}{*}{Majority-vote}      & \multicolumn{1}{c|}{Best} & 81.44   & 82.48     & 81.88    \\
                                    & \multicolumn{1}{c|}{Last} & 80.13   & 79.94     & 79.98    \\ \hline
\multirow{2}{*}{Majority-vote + Ours}      & \multicolumn{1}{c|}{Best} & 82.33    &     84.13 &  \textbf{83.22}   \\
                                    & \multicolumn{1}{c|}{Last} & 79.01   & 79.73    &  79.37    \\ \hline
\end{tabular}
\label{table gleason results}
\end{table}
\subsubsection{Results and Discussion}
The baselines of this study are set to the following:
\begin{itemize}
    \item \textbf{SP} denotes Single Pathologist. We use the annotations provided by one of the pathologists only to train models and calculate the average evaluation results based on six pathologists.  It is noted that the number of training samples for each pathologist is different. So we decide to use the models to make ensemble predictions on the testing set (denoted by \textbf{SP + Ensembles}). 
    \item \textbf{SP + PL}. We leverage Pseudo-Labeling~\cite{lee2013pseudo} technique to train the model using both labeled and unlabeled data from each pathologist. Similarly, we also perform ensemble predictions (\textbf{SP + PL + Ensembles}). 
    \item \textbf{Annotator confusion estimation}~\cite{tanno2019learning} estimates the labeling patterns of the annotators without direct aggregation. We follow \cite{karimi2020deep} and evaluate it on Gleason 2019 dataset for a comparison study.
    \item \textbf{Majority-vote} denotes training from the majority-vote dataset. To further extend the baseline, \textbf{SP + SO} denotes that we use the majority-vote label from the other five pathologists to the unlabeled data, which is similar to the Second Opinion in the clinical practice~\cite{raghu2019direct}.
\end{itemize}
First, we present the results trained from single pathologist (SP) annotations in Fig.~\ref{fig_sp}, aiming to demonstrate a trade-off finding between (1) the ability of a single pathologist to give a correct diagnosis, which denotes the potential noisy rate in individual SP; (2) the number of labeled samples for training; 
(3) the performance of the model. We can observe that although the amount of training data with annotations from pathologist 1 is three times as many as that of pathologist 6, they achieve similar performance. This may be due to the annotations from pathologist 6 has a higher kappa score, which means less label noise existed.
A sub-conclusion can be summarized that the quality of annotations contributes more than the number of labeled samples for training.

Table~\ref{table gleason results} gives the results of the comparative study. Macro recall/precision/F1 score are used as metrics for imbalanced test data.
The semi-supervised technique Pseudo-Labeling~\cite{lee2013pseudo} leverages the unlabeled samples and has made a significantly improved margin to the performance on SP.  
The baselines with ``Ensemble" is able to simulate the clinical decision distribution. It is noticed that Ensemble Model (82.40\% and 81.88\% on the best F1 Score) outperforms direct training from Majority-vote (81.88\% on the best F1 Score) although the number of training samples is the same. Annotator confusion estimation obtains similar results to ensemble models and learns more stably with minimal differences in best and last epochs.
We draw the conclusion that the network implicitly learns information from each pathologist during the training phase and has lower uncertainty on the ensemble predictions, where the uncertainty majority-vote is higher when constructing the ground truth for each sample before training.
Our proposed methods are able to reduce this uncertainty level for the single-target (majority-vote) label and thus improve the performance from 81.88\% to 83.22\% in terms of F1 Score.



\subsection{Kaggle DR+}
\subsubsection{Data Statistics and Comparison}


The original Kaggle Diabetic Retinopathy (DR) dataset~\cite{kagglediabetic} consists of 88,702 fundus photographs from 44,351 patients: one photograph per eye. However, it is estimated that there are about 30\% - 40 \% noisy labels in the originally released dataset~\cite{kagglediabetic}. Moreover, only DR grading labels are included, however, according to~\cite{wang2019retinal}, there are more than 10 retinal diseases such as glaucoma, drusen, et al have never been labelled for this dataset.
Therefore, we decided to re-engineer this dataset and release the multi-label Kaggle DR+ dataset with a golden standard dataset for an unbiased evaluation, which were relabeled by 10+ ophthalmologists covering 17 different retinal diseases commonly examined during screening\footnote{Please refer to our Appendix for a detailed analysis on original and our released dataset}. 
To simplify the noisy label evaluation, we multi-label samples are not considered, and only samples containing sole disease label are used for training and testing. The statistics of selected samples is shown in Table~\ref{Table_kaggle}. 

\subsubsection{Results Analysis}
Table~\ref{table dr} shows the results of two baselines and our proposed method. 
However, it is found that direct training from the original-labeled dataset shows a catastrophic performance due to highly imbalanced data distribution. 
To make a fair comparison, we apply a re-sampling strategy and train baselines from the original-labeled dataset and majority-vote dataset respectively. 
Our proposed method show superior performance compared to the two baselines with re-sampling strategy. 

\begin{figure}[t]
	\includegraphics[width=9cm]{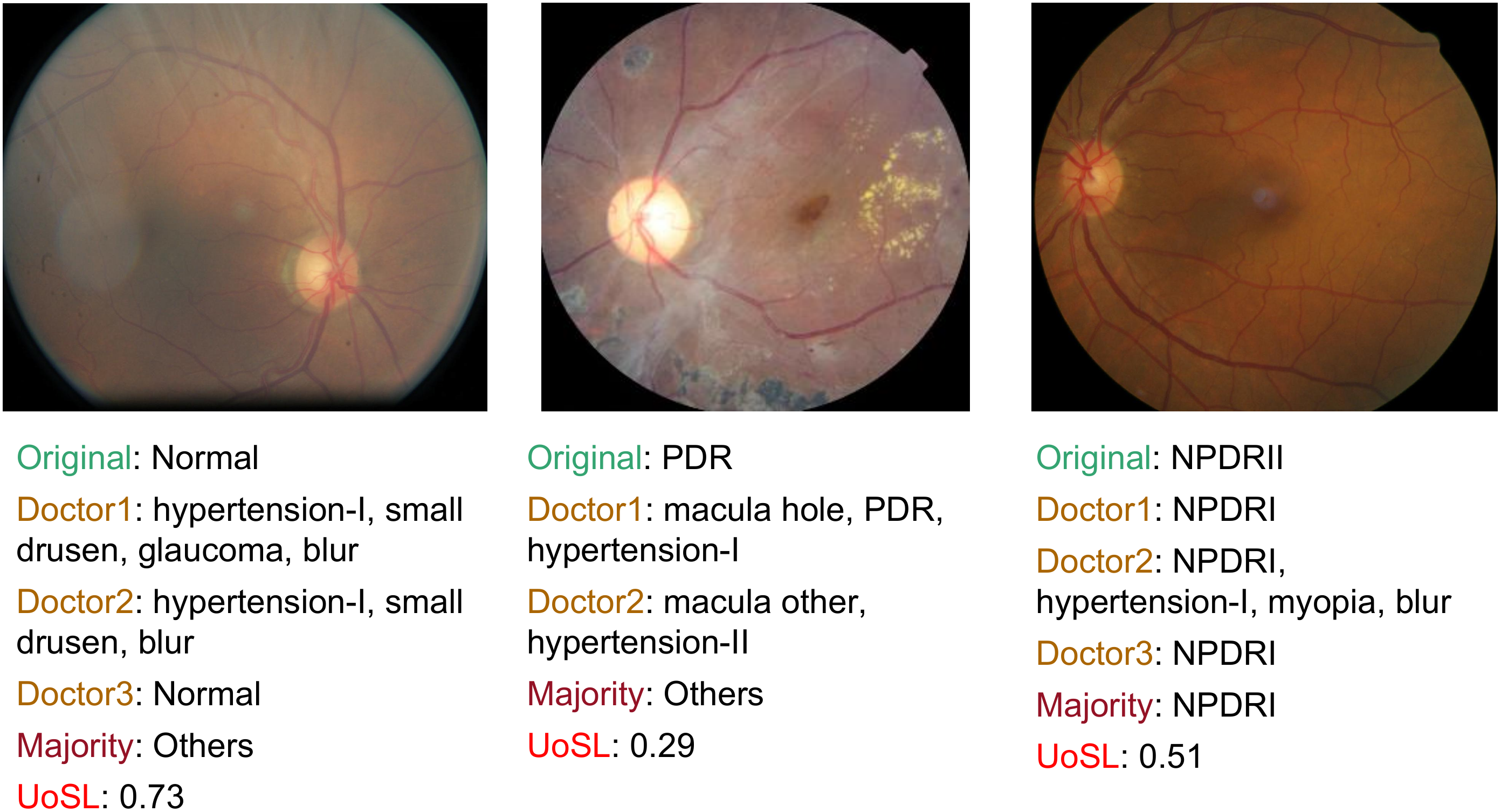}
	\centering
\caption{Examples of applying $UoSL$ for outliers detection. \textbf{Orinigal} defines the DR label from the original Kaggle DR dataset. \textbf{Doctor x} denotes the new label from our Kaggle+ dataset. 
} \label{fig_u_fundus}
\end{figure}

We also give qualitative results in Fig.~\ref{fig_u_fundus} to show the ability of our proposed \emph{UoSL} for outliers (noisy and out-of-category label) detection. 
The first image from the right is wrongly labeled as Normal. The obtained \emph{UoSL} score is a relatively high value of 0.73, which demonstrates our uncertainty-based estimation is sensitive to false-negative samples. 
We also give two bad cases from our method, which explains the limitations of the dual-uncertainty estimation. 
The \emph{UoSL} score is 0.29 on the middle image from Fig.~\ref{fig_u_fundus}. The reason behind may due to the high similarity between PDR (proliferative diabetic retinopathy) and majority-vote results (others, macula and hypertension in this example) in the feature space. 
Also, we find out that \emph{UoSL} is less generalized to mislabelling between two neighbour classes with small appearance differences from the rightmost exmaple (NPDRI and NPDRII, and \emph{UoSL} = 0.51 in the third image).
\subsection{Ablation Study}
To test how each factor makes our model performant, we conduct further ablation studies on various method components. The results of the ablation study indicate that both Focal Loss and Weighted CE can benefit the model compared to CE loss, but Focal Loss with Dual-uncertainty obtain marginal improvements due to the mistaken elimination of some hard samples but with clean labels.

Furthermore, we evaluate the newly-designed iUOD on Gleason 2019 and Kaggle+, which contain annotations from multiple doctors, and the results are shown in Table~\ref{table_ablation}. When $\eta = 0$, the original DUP is applied. We tried several index values for the hyper-parameter $\eta$ to amplify or reduce the effect of a factor which denotes how much attention should be paid to the uncertainty brought by the number of annotations. When $\eta = 1$, the improved DUP achieves the best results evaluated on Gleason 2019. However, there are no more improvements on selected Kaggle DR+. The selected Kaggle DR+ dataset in our manuscript only contains sole disease label and the inconsistency between ophthalmologists are less compared to Gleason 2019.  We can draw a conclusion that improved DUP performs better on the dataset with low quality of annotations and high inconsistency in the number of doctors to give annotations.

\begin{table}[t]
\scriptsize
\centering
\renewcommand\arraystretch{1.2}
\caption{The data statistics on selected samples of Kaggle DR+. {0: Normal; 1: NPDRI; 2: NPDRII; 3: NPDRIII; 4: PDR; 5: Others}.}
\begin{tabular}{lcccccc}
\hline
                & 0 & 1 & 2 & 3 & 4 & 5 \\ \hline
Original        & 8001   & 2978  & 6866   & 1142    & 888 & -      \\
Majority-vote   & 11739  & 1068  & 3359   & 429     & 190 & 3036   \\
Golden Standard & 500    & 200   & 200    & 129     & 123 & -      \\ \hline
\end{tabular}
\label{Table_kaggle}
\end{table}

\begin{table}[t]
\caption{The comparative results of DR classification.}
\centering
\scriptsize
\begin{tabular}{lccc}
\hline
                  & Recall & Precision & F1 Score \\ \hline
Original + Re-sampling & 46.42  & 45.43     & 45.42    \\
Majority-vote + Re-sampling     & 48.51  & 61.30     & 51.05    \\
Ours              & \textbf{49.63}  & \textbf{64.02}     & \textbf{55.91}    \\ \hline
\end{tabular}
\label{table dr}
\end{table}

\vspace{-6pt}

\begin{table}[t]
\centering
\scriptsize
\caption{Ablation study results.}
\begin{tabular}{l|ccccc}
\cline{1-2} \cline{4-6}
\textbf{Component}     & \textbf{ISIC}  &  & \multicolumn{1}{c|}{$\eta$}  & \multicolumn{1}{l|}{\textbf{Gleason}} & \multicolumn{1}{l}{\textbf{Kaggle}} \\ \cline{1-2} \cline{4-6} 
Cross-Entropy & 78.27 &  & \multicolumn{1}{c|}{0 (DUP)} & \multicolumn{1}{c|}{82.53}        & \textbf{55.91}                          \\ \cline{1-2} \cline{4-6} 
Focal Loss    & 80.09 &  & \multicolumn{1}{c|}{1/2}     & \multicolumn{1}{c|}{80.17}        & 55.02                          \\ \cline{1-2} \cline{4-6} 
Dual-U + FL   & 78.77 &  & \multicolumn{1}{c|}{1}       & \multicolumn{1}{c|}{\textbf{83.22}}        & 51.09                          \\ \cline{1-2} \cline{4-6} 
Weighted CE   & 78.92 &  & \multicolumn{1}{c|}{2}       & \multicolumn{1}{c|}{78.26}        & 53.12                          \\ \cline{1-2} \cline{4-6} 
Ours          & 81.01 &  &                              & \multicolumn{1}{l}{}              & \multicolumn{1}{l}{}           \\ \cline{1-2}
\end{tabular} \label{table_ablation}
\end{table}

\section{Conclusions}
\label{Sec. conclusions}
In this paper, we defined and explored two unique types of label noise: disagreement and single-target label noise, existed in the medical image classification task and proposed a Dual-uncertainty estimation framework which can significantly improve model's tolerance to the label noise. 
We conducted extensive experiments on three different disease classification tasks (dermatology, ophthalmology and pathology) and showed the superiority of our proposed framework to handle the uncertainty of annotations in medical images. 
However, our work has some potential limitations since some typical factors in the medical application such as long-tailed and open-label recognition are still needed to been taken into consideration in the presence of label noise. 
We will release an annotation-database which relabeled a public fundus dataset with annotations covering 17 retinal diseases from more than 10 ophthalmologists.  We hope this dataset would help address the  challenges  of label  noise  in  the  medical image analysis.

\bibliographystyle{IEEEtran}
\bibliography{egbib}

\end{document}